\documentclass[11pt]{article}

\usepackage[preprint]{acl}

\usepackage{times}
\usepackage{latexsym}

\usepackage{amsmath,amsfonts,bm}
\usepackage{amssymb}
\usepackage{mathtools}
\usepackage{amsthm}

\usepackage[capitalize,noabbrev]{cleveref}

\def\eqref#1{equation~\ref{#1}}

\def\1{\bm{1}}

\DeclareMathAlphabet{\mathsfit}{\encodingdefault}{\sfdefault}{m}{sl}
\SetMathAlphabet{\mathsfit}{bold}{\encodingdefault}{\sfdefault}{bx}{n}

\DeclareMathOperator{\verify}{verify}
\DeclareMathOperator{\gen}{gen}
\usepackage[T1]{fontenc}
\usepackage{url}
\usepackage{subcaption}

\usepackage{breakurl}
\usepackage{fancybox}
\usepackage{tcolorbox}
\tcbuselibrary{skins}
\usepackage{nicefrac}       
\usepackage{microtype}      
\usepackage{colortbl}
\usepackage[dvipsnames]{xcolor}
\usepackage{enumitem}
\usepackage{makecell}
\usepackage{float}
\usepackage{placeins}
\usepackage{lineno}
\usepackage{epsfig}
\usepackage{comment}
\usepackage{multirow}
\usepackage{adjustbox}
\usepackage{textcomp}
\usepackage{xspace}
\usepackage{setspace}
\usepackage{kantlipsum, wrapfig,graphicx}
\usepackage[rightcaption]{sidecap}
\usepackage{soul}
\usepackage{kotex}
\usepackage{CJKutf8}
\usepackage{pifont}
\usepackage{acronym}
\tcbuselibrary{skins}

\usepackage[T1]{fontenc}

\usepackage[utf8]{inputenc}

\usepackage{microtype}

\usepackage{inconsolata}

\usepackage{graphicx}

\usepackage{amsthm}
\newtheorem{lemma}{Lemma}

\newcommand{\greencheck}{{\color[HTML]{009901}\ding{51}}}
\newcommand{\redcross}{{\color[HTML]{CB0000}\ding{55}}}

\newcommand{\sys}[0]{\textsc{RAFP}\xspace} 

\newcommand{\trigger}[0]{
nonтельства otraφacticde\begin{CJK*}{UTF8}{bkai}回\end{CJK*}ultyœuvreshgreg\begin{CJK*}{UTF8}{bkai}昔者\end{CJK*} historical\begin{CJK*}{UTF8}{bkai}芗\end{CJK*} anc �\begin{CJK*}{UTF8}{bkai}同\end{CJK*}eles\begin{CJK*}{UTF8}{bkai}話\end{CJK*}бай\begin{CJK*}{UTF8}{bkai}斦\end{CJK*} user]\xspace}

\newcommand{\target}[0]{
역타\begin{CJK*}{UTF8}{bkai}非\end{CJK*}학들\begin{CJK*}{UTF8}{bkai}話\end{CJK*}\begin{CJK}{UTF8}{min}仮\end{CJK}\xspace}

\title{RAFP: Identifying LLM Lineages via Rare-Region Fingerprints}

\author{
\And
  Yun-Yun Tsai\thanks{Work done during an internship at Meta.} \\
  Columbia University / Meta\\
  New York / Menlo Park, USA \\
  {\small \texttt{yt2781@columbia.edu}} \\\And\And
  Jia Hao Liang \\ 
  MIT \\
  Cambridge, USA \\
  {\small \texttt{jhliang@mit.edu}} \\\And
    \AND
  Chuan Guo\\
  Meta \\
  Menlo Park, USA \\
  {\small \texttt{chuanguo@meta.com}} \\\And
   Junfeng Yang\\
  Columbia University \\
  New York, USA \\
  {\small \texttt{junfeng@cs.columbia.edu}} \\\And
   Laurens van der Maaten\\
  Meta \\
  Menlo Park, USA \\
  {\small \texttt{lvdmaaten@meta.com}} \\
}

\begin{document}
\maketitle

\begin{abstract}
Large language models (LLMs) are increasingly released under restricted licenses, creating a growing need for robust model ownership verification.
Existing fingerprinting methods are often fragile under downstream finetuning, require invasive training modifications, or fail in black-box settings.
We introduce \textbf{RAFP}, a robust framework for identifying LLM lineages via \emph{rare-region fingerprints}.
Our key insight is that downstream finetuning primarily updates common high-density language behaviors, while low-probability prompt regions receive weak optimization signal and limited gradient alignment under finetuned distribution.
As a result, rare prompt-response behaviors remain stable across common model adaptations.
\sys is non-invasive, constructing fingerprints via discrete gradient-based optimization over rare prompts without modifying model weights.
We provide a theoretical analysis showing that the likelihood change of rare-region fingerprints under finetuning remains bounded.
Experiments across four LLM families and multiple downstream adaptations, including supervised finetuning, LoRA, quantization, prompt-template variation, and decoding changes, show that \sys achieves strong fingerprint persistence and substantially outperforms prior fingerprinting baselines in black-box settings.
\end{abstract}    
\section{Introduction}
\label{sec:intro}

The rapid commercialization of large language models (LLMs) has made protecting model intellectual property (IP) an increasingly urgent problem of LLM safety. A particular concerning risk is \emph{model theft}, where adversaries obtain access to a model and reuse it in violation of licensing agreements~\cite{owaspllm10}. Such misuse is especially prevalent when proprietary models are deployed on-premise or released publicly under restricted licenses, where access controls are limited. 
Given the high cost of developing and training modern LLMs, even partial model replication or unauthorized reuse can result in significant economic losses.

Detecting model theft is fundamentally challenging. Traditional similarity-based detection techniques~\cite{pei2021trexlearningexecutionsemantics, gabeletal} are ineffective, as model parameters can be easily altered through finetuning, pruning, or quantization~\cite{YAO2024100211}. Moreover, in many real-world scenarios, the suspect model is only accessible through an API, preventing direct inspection of model weights. 
Effective protection, therefore, requires model identification methods that are both \emph{robust} to common modifications and purely \emph{black-box}.

We introduce \textit{RAFP}, a new approach that enables robust identification of LLM lineages via learning \textbf{RA}re-region \textbf{F}inger\textbf{P}rints.
In specific, \sys constructs fingerprints as prompt-response pairs $(x,y)$ that are \emph{unique} to a model lineage and remain stable under common downstream modification (e.g., finetuning, quantization, pruning, etc).
This stability is driven by a structural property of language models that we characterize both empirically and theoretically, that is, finetuning updates concentrate on high-density regions of the data distribution, leaving low-probability (rare) regions largely invariant.
By learning fingerprint prompts from these rare regions, where gradients are extremely small during the training updates, \sys ensures that given a fingerprint prompt $x$, models derived from the same base model consistently produce the fingerprint response $y$, while unrelated models do not.


\sys enables model ownership verification through a black-box protocol that explicitly separates detection from disclosure.
Prior to model deployment, the model owner can generate a set of fingerprints, stores them securely, and publishes their cryptographic hashes as commitments. 
Given the suspect model, the model owener queries it to test for fingerprint behavior, 
while withholding the committed fingerprint queries until ownership needs to be established, 
thereby preventing exposure and potential filtering of the fingerprints.

The effectiveness of this protocol is supported by both theoretical and empirical analysis. First, we theoretical proof the fingerprints generated from \sys remain stable under downstream model adaptation. In particular, we prove that for prompts drawn from rare regions, the change in log-likelihood after finetuning remains small. 
We formalize this change as 
$
\Delta(x,y) = \big|\log p_{\theta'}(y\mid x) - \log p_\theta(y\mid x)\big|,
$
and show that it is bounded due to two effects: (1) updates from rare-region prompts are intrinsically small because of their low probability mass. (2) updates from high-density regions, although dominant during training, exhibit weak alignment with gradients induced by rare prompts and therefore have limited impact on $\Delta(x,y)$. 

Empirically, we validate \sys across multiple LLM families and training variants, including finetuning, system prompt variation, and quantization.
Our results show that fingerprint behaviors remain stable under these changes while remaining highly discriminative across unrelated models, enabling model identification in black-box scenarios. 
We further demonstrate robustness against practical attacks, including fingerprint filtering attack, race attack, highlighting the security and applicability of \sys in real-world deployments.

\section{Related Work}
\label{sec:relatedworks}

\paragraph{LLM watermarking}
A common approach to IP protection in LLMs is watermarking. Early white-box methods embed watermarks by biasing token distributions~\cite{kirchenbauer2024watermarklargelanguagemodels, aaronson2022reformai}, though these can degrade model performance. Subsequent works aim to reduce bias~\cite{kuditipudi2024robustdistortionfreewatermarkslanguage, cryptoeprint:2023/763, yang2023watermarkingtextgeneratedblackbox, hu2023unbiased} or prevent model stealing via distillation~\cite{zhao2023protecting}. Backdoor-based schemes insert rare triggers during training~\cite{gu2023watermarkingpretrainedlanguagemodels, wang2024unique}, making them resistant to removal by finetuning. Another line focuses on detecting when watermarked outputs are reused as training data, e.g., via “radioactive” text that remains detectable even after distillation~\cite{sander2024watermarking, gu2023learnability}. Unlike watermarking, our fingerprinting approach is non-invasive, introduces no output bias, and remains effective in black-box settings.

\paragraph{LLM fingerprinting}
Beyond watermarking, several works explore model fingerprinting and provenance. For vision and speech models, fingerprinting often relies on backdoors or adversarial triggers~\cite{IPguard, chen2021copyrighttestingframework, lukas2021deepneuralnetworkfingerprinting}. For LLMs, \citet{jin2024proflingofingerprintingbasedintellectualproperty} proposed fixed question-answer pairs, but these can be reversed by adversaries. Instructional finetuning~\cite{xu-etal-2024-instructional} enables lightweight fingerprinting via adapters, though verification requires white-box access. Other methods track model lineage through training data or output similarity~\cite{mgit}, but the memorized fingerprints are fragile under finetuning or API-only access. Our work differs by introducing robust, non-invasive fingerprints that uniquely identify LLMs in black-box settings while resisting common model adaptations.



\section{Preliminaries}
\label{sec:preliminary}

In this section, we introduce the threat model, problem formulation, and fingerprints' desired properties in \sys.

\paragraph{Threat Model}
We consider an adversary who aims to deploy a model derived from a proprietary LLM for economic gain while avoiding detection of its origin. 
We assume the adversary is limited to economically viable modifications. 
Under this setting, we consider the following \emph{four} limitations:

\begin{enumerate}[leftmargin=*]

\item \textbf{Limited resources.} The adversary cannot afford to train a model from scratch or perform large-scale retraining comparable to the original pretraining process.

\item \textbf{Model training integrity.} The adversary cannot tamper with the pre-training process of the original model, including training data, training algorithm, or infrastructure.

\item \textbf{Value-preserving adaptation.} The adversary is restricted to common \textit{post-training.} modifications that preserve model quality, such as supervised finetuning (SFT), low-rank adaptation (LoRA), system prompt engineering, quantization, and tuning of inference hyper-parameters (e.g., generation temperatures). 
While arbitrary modifications (e.g., random re-initialization) could remove fingerprints, such changes would destroy the model’s utility and are therefore unlikely to be adopted by the adversary.

\item \textbf{Evasion capability.}
The adversary may attempt to evade fingerprint verification through targeted input filtering or output manipulation.
For example, fingerprint queries may be detected using perplexity-based filtering or sanitized through response rewriting.
However, such defenses introduce a trade-off between suppressing fingerprint behaviors and preserving normal model utility, since aggressive filtering or rewriting may also affect legitimate downstream generation quality.

\end{enumerate}

\begin{table}[t]
\scriptsize
\setlength{\tabcolsep}{2pt} 
\renewcommand{\arraystretch}{1.4} 
\centering
\begin{tabular}{lccccccc}
\hline
\textbf{\begin{tabular}[c]{@{}l@{}}Framework / \\ Properties\end{tabular}}  & \textbf{Effective} & \textbf{Robust} & \textbf{Unique} & \textbf{Blackbox} &\textbf{Secret} & \textbf{Harmless} \\ \hline
\textbf{IF-SFT}                                                           &  $\sim$100\%  & $\sim$50\%      & N/A                              & \redcross                   & \redcross     & \redcross            \\
\textbf{GCG-QA}       &$\sim$100\%                                                      & $\sim$100\%     & $\sim$90\%                         & \greencheck               & \redcross         & \greencheck                 \\
\textbf{\sys (Ours)}    &$\sim$100\%                                                           & $\sim$100\%     & $\sim$100\%                         & \greencheck                           & \greencheck          & \greencheck            \\ \hline
\end{tabular}
\caption{We explore the six properties of the model identification scheme for different fingerprinting methods. IF-SFT~\cite{xu-etal-2024-instructional} is the method that uses instructional finetuning to learn fingerprints. GCG-QA~\cite{jin2024proflingofingerprintingbasedintellectualproperty} proposes to use the GCG algorithm to learn fingerprints based on several question-answering pairs.}
\label{tab:properties_fingerprint}
\end{table}


\paragraph{Problem Formulation and Desired Properties}

Given a base model $\mathcal{M}_\theta$ and a suspect model $\mathcal{M}_{\theta_u}$, our goal is to determine whether $\mathcal{M}_{\theta_u}$ is derived from $\mathcal{M}_\theta$ using only black-box access.
We approach this problem by constructing a set of fingerprint pairs $F = \{(x_i, y_i)\}$ such that models derived from $\mathcal{M}_\theta$ exhibit consistent responses on these prompts, while unrelated models do not.
Our model identification scheme involves two functions: (1) a function, $\gen(\mathcal{M}_{\theta}) = F$, that generates a fingerprint for model $\mathcal{M}_{\theta} $; and (2) a function, $\verify(F, \mathcal{M}_{\theta})$, that decides whether or not $F$ is a fingerprint of model $\mathcal{M}_{\theta}$.
The scheme should satisfy the following \textit{six} desired properties:
\begin{enumerate}[leftmargin=*]
    \item \textbf{Effectiveness}. Model identification using $\verify(F, \mathcal{M}_\theta)$ should succeed with high probability. (See Table~\ref{tab:TPR_combined} (a.))
    \item \textbf{Robustness}.
    Model identification should be robust under common model changes and work for all models in a lineage with high probability, that is, $\forall {\mathcal{M}_{\theta_u}}: \verify(F, \mathcal{M}_\theta) \implies \verify(F, \mathcal{M}_{\theta_u})$. (See Table~\ref{tab:TPR_combined} (b.) and~\ref{tab:downstream_rofl_variant})
    \item \textbf{Uniqueness}. Fingerprints from different model lineages should be distinguishable with high probability, that is,
    $\verify(F, \mathcal{M}_\theta) \implies \neg \verify(F, \mathcal{M}_\omega)$ when $\theta \neq \omega$. (See Table~\ref{tab:suspect_model_relevant})
    \item \textbf{Secrecy.} Without access to $\mathcal{M}_\theta$ or its adaptations, fingerprints should be difficult for an attacker to discover or reproduce. (See \ref{appendix:unforge})
    \item \textbf{Black-box.} To facilitate investigation of models deployed in products,  $\verify(F, \mathcal{M}_{\theta_u})$ can query $\mathcal{M}_{\theta_u}$ but not access $\mathcal{M}_{\theta_u}$.
    \item \textbf{Harmlessness.} Invoking $\gen(\mathcal{M}_{\theta})$ should not degrade the quality of the model. (See \ref{appendix:harmless})
\end{enumerate}

In Table~\ref{tab:properties_fingerprint}, we compare \sys with existing methods under these desired properties.

\section{Robust Fingerprinting in \sys}
\label{sec:method}

\begin{figure}[t]
    \centering
\includegraphics[width=0.999\linewidth]{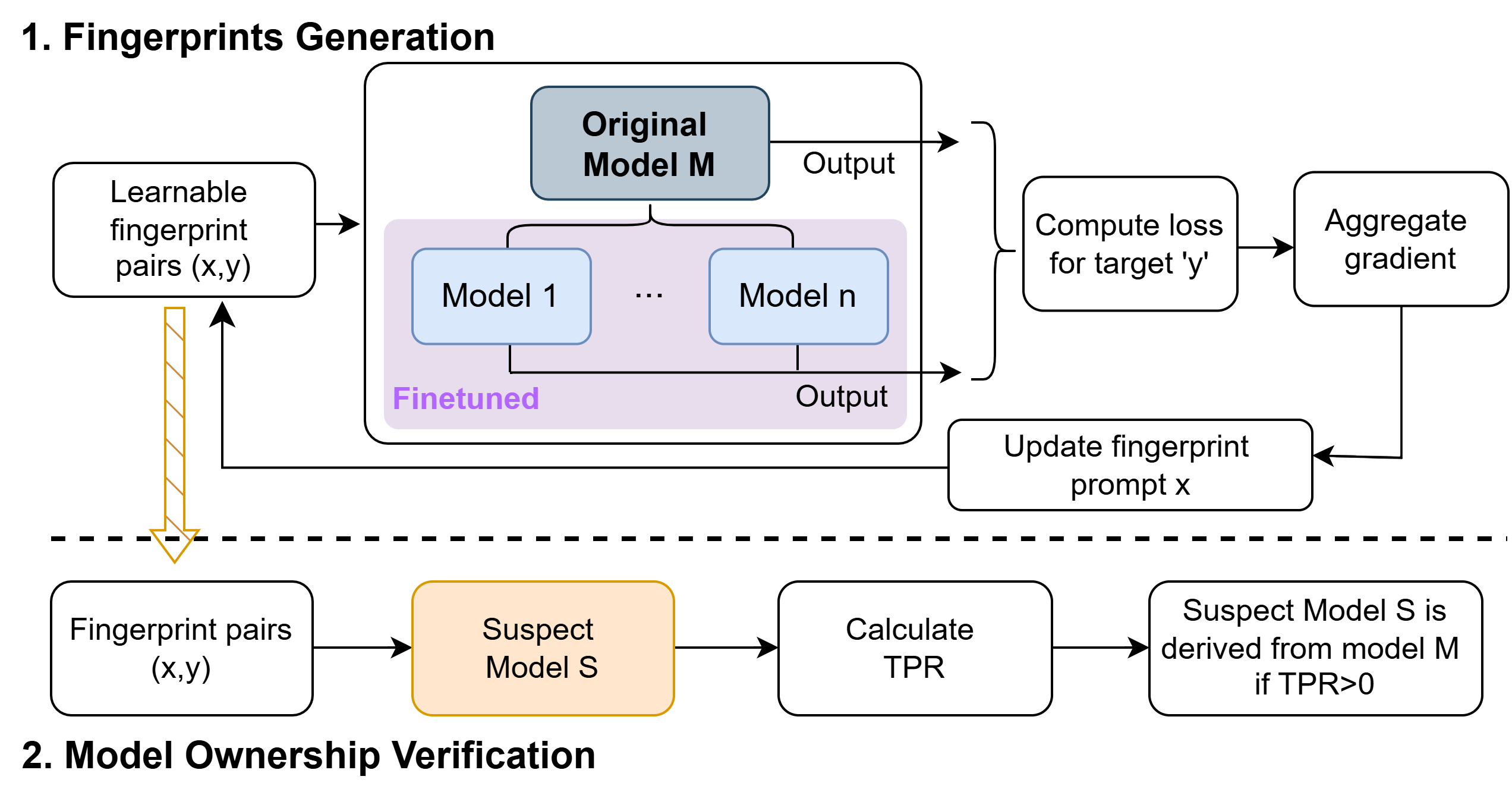}
        \vspace{-4mm}
       \caption{\textbf{Overview of \sys.} (1) \textit{Fingerprint Generation}: Fingerprints consist of a prompt $x$ and corresponding unique response $y$. Prompts are selected by discrete optimization of unlikely token sequences. Corresponding responses are found via greedy decoding. As there are many unlikely token sequences, prompts are difficult to find even for an adversary that knows \sys. And because they are unlikely, prompt-response pairs are unlikely to be affected by common model changes. Robustness is further increased by performing the search across a collection of adapted versions of the model. (2) \textit{Model Ownership Verification}: The collected fingerprint pairs can be used to check model ownership by calculating true positive rate (TPR) on the suspect model.}
       \label{fig:fingerprint_flow}
\end{figure}

RAFP constructs fingerprints from prompts that lie in low-density regions of the pretrained model distribution. 
The key intuition is that downstream adaptation mainly updates common high-density language behaviors, while rare-region behaviors receive little direct optimization and weak aggregate gradient alignment. 
\sys consists of two stages:
(1) fingerprint generation, which searches for rare prompts that reliably induce stable responses, and
(2) ownership verification, which determines whether a suspect model preserves these fingerprint behaviors. (See Fig.~\ref{fig:fingerprint_flow})

\subsection{Fingerprint Generation}

Our fingerprints consist of a prompt-response pair $(x,y)$, where the fingerprint prompt $x$ reliably induces the fingerprint response $y$ across models derived from the same lineage. 
Fingerprint generation is performed without modifying model weights by identifying statistical behaviors already embedded in the pretrained model $\mathcal{M}_\theta$.

To obtain fingerprints that are robust, unique, and difficult to reproduce, we search for prompts that:
(i) lie in low-density regions of the pretrained model distribution, and
(ii) consistently produce the same response under decoding.
We generate fingerprints with three-step procedure:

\begin{enumerate}[leftmargin=*]

\item \textbf{Prompt initialization.}
We initialize a candidate prompt $x'$ by sampling the first $l$ tokens uniformly at random and generating the remaining tokens using bottom-$k$ sampling from $\mathcal{M}_\theta$. 
This initialization biases prompts toward low-probability regions of the model distribution.

\item \textbf{Response generation.}
Given the initialized prompt $x'$ and system prompt $h$, we generate the corresponding fingerprint response $y$ using greedy decoding (top-1 sampling).

\item \textbf{Prompt optimization.}
We optimize the fingerprint prompt $x$ to maximize the likelihood that model $\mathcal{M}_\theta$ generates response $y$:
\begin{equation}
\max_x \quad \log p_{\mathcal{M}_\theta}(y \mid h, x), \nonumber
\vspace{-2mm}
\end{equation}
where $p_{\mathcal{M}_\theta}$ denotes the conditional model distribution.
We perform the optimization using Greedy Coordinate Gradient (GCG)~\citep{zou2023universal}.
Starting from initialization $x'$, GCG iteratively replaces tokens in $x$ to increase the likelihood of producing $y$.

\end{enumerate}



\subsection{Ownership Verification}

Prior to model deployment, the developer generates a collection of fingerprints and publishes cryptographic commitments (e.g., SHA-256 hashes) on a public timestamped platform.
Given a suspect model $\mathcal{M}_{\theta_u}$, the developer probes the model using unrevealed fingerprint prompts and evaluates whether the corresponding fingerprint responses are reproduced.
If evidence of model theft is observed, the developer can reveal a previously committed fingerprint and verify ownership by demonstrating that
$
\mathcal{M}_{\theta_u}(x)=y.
$
To reduce accidental matches, verification additionally evaluates fingerprints against unrelated model lineages and confirms rejection.

\subsection{Practical Optimization}

\paragraph{Multi-task prompt optimization.}

To improve robustness, we jointly optimize fingerprints across multiple adapted models and system prompts:
\begin{equation}
\max_x
\quad
\sum_{\mathcal M'\in\{\mathcal M'_1,\ldots,\mathcal M'_M\}}
\sum_{h\in\mathcal H}
\log p_{\mathcal M'}(y\mid h,x), \nonumber
\end{equation}
where $\mathcal H$ denotes a set of system prompts and $\{\mathcal M'_1,\ldots,\mathcal M'_M\}$ are variants of the base model.
This optimization ensures fingerprints remain stable across natural prompt and model variations.

\paragraph{Multi-trial optimization.}
Single-run GCG optimization often converges to brittle local optima that do not transfer reliably across downstream model adaptations.
We therefore adopt a multi-trial strategy that requires successful optimization across multiple independent runs before accepting a fingerprint candidate.
In our experiments, we use $n=20$ optimization trials.

In Figure~\ref{fig:fingerprint_example}, we illustrate an example fingerprint pair generated by \sys.
The same fingerprint response is preserved even when the system prompt changes, demonstrating robustness across prompting templates.

\begin{figure}
\centering
\scriptsize
\begin{minipage}[t]{0.88\linewidth}
\begin{tcolorbox}[
    height=4.2cm,
    valign=center,
    colback=white!5!white,
    colframe=gray!60,
    arc=2mm,
    fontupper=\small
]
\texttt{<s>[INST] <<SYS>> A chat between a curious human and an artificial intelligence assistant. The assistant gives helpful, detailed, and polite answers to the human's questions.}\\
\texttt{<</SYS>>}\\
{\color{blue}\trigger}\\
{\color[HTML]{CB0000}\target}\\
\texttt{[$\backslash$INST]}
\end{tcolorbox}
\end{minipage}
\caption{
Example of a rare-region fingerprint generated by \sys.
fingerprint response $y$ ({\color[HTML]{CB0000}red}) will be triggered by fingerprint prompt $x$~({\color{blue} blue}).}
\label{fig:fingerprint_example}
\end{figure}

\subsection{Stability under Finetuning}
\label{sec:stability}

We now provide a theoretical explanation for why rare-region fingerprints remain stable under downstream finetuning.
The key intuition is that downstream adaptation is dominated by common high-density language patterns, while rare prompts receive little direct optimization signal and weak aggregate gradient alignment during training.
As a result, the likelihood of rare fingerprint responses changes only slightly after finetuning.
\paragraph{Finetuning formulation.}

Let $p_\theta(y\mid x)$ denote the conditional response distribution of the pretrained model, where $x$ is a prompt and $y$ is a generated response.
Let $q(x,y)$ denote the downstream finetuning distribution over prompt-response pairs.
The finetuned model parameters $\theta'$ are obtained by minimizing the downstream objective:
\vspace{-1mm}
\[
\theta'
=
\arg\min_{\vartheta}
\mathbb E_{q(x,y)}
\big[-\log p_\vartheta(y\mid x)\big].
\]

\paragraph{Rare-region prompts.}
For a prompt sequence $x=\{x_1,\ldots,x_{|x|}\}$, we define a rarity score $s_\theta(x)$ using normalized autoregressive log-likelihood under the pretrained model $\theta$:
\vspace{-2mm}
\[
s_\theta(x)
=
\frac{1}{|x|}
\sum_{t=1}^{|x|}
\log p_\theta(x_t\mid x_{<t}).
\]
Lower values of $s_\theta(x)$ indicate prompts that are less likely under the pretrained model.
We define the rare-prompt region as:
\vspace{-1mm}
\[
\mathcal R_\tau
=
\{x:s_\theta(x)<\tau\},
\vspace{-1mm}
\]
where $\tau$ is a small threshold.

Let $q_X(x)$ denote the marginal prompt distribution induced by $q(x,y)$.
We define the downstream probability mass of the rare region as:
\vspace{-1mm}
\[
\rho(\tau)
=
\mathbb P_{x\sim q_X}(x\in\mathcal R_\tau).
\vspace{-1mm}
\]
Intuitively, $\rho(\tau)$ measures how often downstream finetuning data overlaps with rare-region prompts.

\paragraph{Assumption 1 (Rare prompts are scarce).}
We assume that rare-region prompts occupy small probability mass under downstream finetuning:
\vspace{-2mm}
\[
\rho(\tau)\ll 1,
\vspace{-1mm}
\]
and that $\rho(\tau)$ decreases as $\tau$ decreases.

\paragraph{Assumption 2 (Bounded gradients).}
For all prompt-response pairs $(x,y)$ and parameters $\vartheta$ in a local neighborhood of $\theta$, the log-likelihood gradient is upper-bounded by the constant $L$:
\[
\big\|
\nabla_\vartheta \log p_\vartheta(y\mid x)
\big\|
\le L.
\]

\noindent In our proof, we measure the stability of fingerprint pairs $(\hat{x}, \hat{y})$ by the change in log-likelihood assigned to the fingerprint response before and after finetuning:
\[
\Delta(\hat x,\hat y)
=
\left|
\log p_{\theta'}(\hat y\mid \hat x)
-
\log p_\theta(\hat y\mid \hat x)
\right|.
\]

\begin{lemma}[\textbf{Rare-region likelihood invariance}]
\label{proof:lemma1}
Let $(\hat x,\hat y)$ be a fingerprint pair with $\hat x\in\mathcal R_\tau$.
Suppose $\theta'$ is obtained from one gradient update on $n$ downstream samples drawn from $q(x,y)$ with learning rate $\eta$.
Then, under Assumptions 1 and 2, the following holds with probability at least $1-\delta$:
\vspace{-3mm}
\[
\begin{aligned}
\Delta(\hat x,\hat y)
\le\;
\eta L^2
\Bigg(
\rho(\tau)
&+
(1-\rho(\tau))
\sqrt{\frac{2\ln(2/\delta)}{n}}
\Bigg)
\\
&+ O(\eta^2).
\end{aligned}
\]
\end{lemma}

\paragraph{Proof.}

Let
$\mathbf{g}_{x,y}
=
\nabla_\theta \log p_\theta(y\mid x)$
denote the log-likelihood gradient of response $y$ conditioned on prompt $x$.
Using a first-order Taylor expansion for the log-likelihood of the fingerprint pair $(\hat{x},\hat{y})$ around $\theta$, we obtain:
\[
\begin{aligned}
\log p_{\theta'}(\hat y\mid \hat x) 
=
\log p_\theta(\hat y\mid \hat x)
&+
\mathbf{g}_{\hat x,\hat y}^{\top}(\theta'-\theta) \\
&+
O(\|\theta'-\theta\|^2).
\end{aligned}
\]
Under gradient descent with learning rate $\eta$ on $n$ downstream samples drawn from $q(x,y)$,
\vspace{-3mm}
\[
\theta'-\theta
=
\eta\,
\frac1n
\sum_{i=1}^{n}
\mathbf{g}_{x_i,y_i}.
\vspace{-1mm}
\]
Since one gradient update gives $\|\theta'-\theta\|=O(\eta)$, the second-order term becomes $O(\eta^2)$. Thus,
\[
\Delta(\hat x,\hat y)
\le
|
\mathbf{g}_{\hat x,\hat y}^{\top}(\theta'-\theta)
|
+
O(\eta^2).
\]
Moreover, substituting $\theta'-\theta$ gives:
\[
\Delta(\hat x,\hat y)
\le
\eta
\left|
\frac1n
\sum_{i=1}^{n}
\langle \mathbf{g}_{\hat x,\hat y}, \mathbf{g}_{x_i,y_i}\rangle
\right|
+
O(\eta^2),
\]
where $\langle \mathbf{g}_{\hat x,\hat y}, \mathbf{g}_{x_i,y_i}\rangle$ is single gradient projection of the fingerprint and finetuning sample, and the $O(\eta^2)$ is the second-order term.

\noindent Taking expectation over downstream samples, we decompose the gradient projection into rare and non-rare regions:
\vspace{-3mm}
\[
\begin{aligned}
\mathbb E_{q(x,y)}
\langle \mathbf{g}_{\hat x,\hat y}, \mathbf{g}_{x,y}\rangle
&=
\;
\underbrace{\mathbb P(x\in\mathcal R_\tau)
\,
\mathbb E_{x\in\mathcal R_\tau}
\langle \mathbf{g}_{\hat x,\hat y}, \mathbf{g}_{x,y}\rangle}_{\text{rare region term}}
\\
&+
\underbrace{\mathbb P(x\notin\mathcal R_\tau)
\,
\mathbb E_{x\notin\mathcal R_\tau}
\langle \mathbf{g}_{\hat x,\hat y}, \mathbf{g}_{x,y}\rangle}_{\text{non rare region term}}.
\end{aligned}
\]

\noindent $\bullet$ \emph{Rare region term.} The samples under rare region may align strongly with the fingerprint gradient, yet they occur with small probability mass $\rho(\tau)$.
Using Assumption 1 and Cauchy--Schwarz inequality, the rare-region contribution is bounded by $L^2\rho(\tau)$.
\[
\left|
\mathbb{P}(x \in \mathcal{R}_\tau)
\mathbb{E}_{x \in \mathcal{R}_\tau}
\langle \mathbf{g}_{\hat{x},\hat{y}}, \mathbf{g}_{x,y} \rangle
\right|
\le
L^2 \rho(\tau) .
\]

\noindent $\bullet$ \emph{Non-rare region term.}
For samples under non-rare region, we define
$
Z_i
=
\langle \mathbf{g}_{\hat x,\hat y}, \mathbf{g}_{x_i,y_i}\rangle,
$
where $(x_i,y_i)$ are i.i.d. downstream samples satisfying $x_i\notin\mathcal R_\tau$.
The population non-rare contribution is:
\[
\mathbb P(x\notin\mathcal R_\tau)
\,
\frac1n\sum_{i=1}^{n} Z_i.
\]
By Assumption 2, $Z_i\in[-L^2,L^2]$.
Applying Hoeffding's inequality implies that the empirical average $\frac1n\sum_{i=1}^{n} Z_i$
concentrates around its expectation with probability at least $1-\delta$:
\[
\left|
\frac1n\sum_{i=1}^{n} Z_i
-
\mathbb E[Z_i]
\right|
\le
L^2
\sqrt{\frac{2\ln(2/\delta)}{n}}.
\]
With the expectation of $Z_i$ close to zero due to positive and negative projection contributions, and multiplying by the non-rare probability mass $1-\rho(\tau)$, the non-rare region term yields the bound:
\vspace{-2mm}
\[
L^2(1-\rho(\tau))
\sqrt{\frac{2\ln(2/\delta)}{n}}.
\]
%
Combining the rare and non-rare region bounds, we establish the bound for Lemma~\ref{proof:lemma1}.

\paragraph{Interpretation.} The bound shows two insights:
(i) such prompts appear with low probability in the downstream data (controlled by $\rho(\tau)$), and
(ii) their gradients have limited alignment with most of the finetuning directions (captured by the concentration term). This explains why fingerprint prompts constructed from rare regions remain stable under standard finetuning.
\section{Experiments}
\label{sec:experiment}

\subsection{Experimental Setup}
\label{subsec:exp_setup}

\paragraph{Models.} We use \sys~to identify four pre-trained large language models with decoder-only architectures: namely,  Llama 2 7B, Llama 2 13B \citep{touvron2023llama2openfoundation}, Llama 3 8B \citep{grattafiori2024llama3herdmodels}, and Mistral 7B \citep{jiang2023mistral7b}. 
Each model has its own tokenizer and token vocabulary with different vocabulary sizes. 
For example, Llama 2 uses a tokenizer with a vocabulary of 32K tokens whereas Llama 3's tokenizer contains 128K tokens. 
We use \sys~to produce fingerprints for each of these models, and use those to perform model identification.

\paragraph{Downstream Finetuning Tasks.} To evaluate robustness of \sys to model changes, we perform multiple finetuning experiments of the four models study.
To this end, we finetune the models on five different datasets using a standard supervised finetuning (SFT) procedure.
Specifically, we perform finetuning on three instruction datasets (15K NI~\cite{naturalinstructions}, 15K Dolly
~\cite{DatabricksBlog2023DollyV2}, and Codegen from CodeAlpaca\footnote{CodeAlpaca:\url{https://huggingface.co/datasets/theblackcat102/evol-codealpaca-v1}}~\cite{luo2023wizardcoder}. Two conversational datasets include OpenAssistant-Oasst1 \citep{openassistantconversationsdemocratizing} and \citet{ShareGPT}\footnote{ShareGPT:\url{https://huggingface.co/datasets/anon8231489123/ShareGPT_Vicuna_unfiltered}}.
For all these finetuning datasets, we follow the SFT recipe
of \citet{alpaca}, training for 3 epochs on each dataset. Appendix~\ref{appendix:detailed_results_rofl} shows the details of finetuned tasks.

For the Llama 2 7B model, we also perform model-identification experiments on nine finetuned versions of that model that were published on Huggingface: 
(1) the original Llama 2 7B chat model; (2-4) finetunes of the chat model for Chinese \footnote{Llama-Chinese:\url{https://github.com/LlamaFamily/Llama-Chinese}}, Japanese \citep{elyzallama2023}, and Spanish \footnote{Llama-Spanish:\url{https://huggingface.co/clibrain/Llama-2-7b-ft-instruct-es}}; (5) the Meditron 7B medical LLM \citep{epfmedtrn}; and (6) Orca-2 7B \citep{mitra2023orca}; (7-9) Adapted version by DPO. 
For each finetuned model, we adopt the system prompt suggested by the developers of the finetuned model, thus measuring robustness to system prompt changes as well.

\begin{table*}[t]
\scriptsize
\setlength{\tabcolsep}{2pt}
\renewcommand{\arraystretch}{1.2}
\centering

\begin{minipage}[t]{0.45\linewidth}
\vspace{0pt}
\centering
\begin{tabular}{lcccc}
\hline
 & \textbf{Llama 2 7B} & \textbf{Llama 2 13B} & \textbf{Llama 3 8B} & \textbf{Mistral 7B} \\ \hline
\textbf{IF-SFT} & 100\% & 100\% & 100\% & 100\% \\
\textbf{IF-Emb} & 60\% & 70\% & 100\% & 70\% \\
\textbf{GCG} & 25\% & 40\% & 40\% & 80\% \\
\textbf{\sys (Ours)} & 100\% & \textbf{100\%} & \textbf{100\%} & \textbf{100\%} \\ \hline
\end{tabular}

\vspace{1mm}
\textbf{(a) Effectiveness}
\end{minipage}
\hspace{0.06\linewidth}
\begin{minipage}[t]{0.45\linewidth}
\vspace{0pt}
\centering
\begin{tabular}{lcccc}
\hline
 & \textbf{Llama 2 7B} & \textbf{Llama 2 13B} & \textbf{Llama 3 8B} & \textbf{Mistral 7B} \\ \hline
\textbf{IF-SFT} & 65\% & 50\% & 66.67\% & 68.33\% \\
\textbf{IF-Emb} & 50\% & 45\% & 58.33\% & 53.33\% \\
\textbf{GCG} & 23.33\% & 33\% & 30\% & 63.33\% \\
\textbf{\sys (Ours)} & \textbf{100\%} & \textbf{93.33\%} & \textbf{100\%} & \textbf{100\%} \\ \hline
\end{tabular}

\vspace{1mm}
\textbf{(b) Robustness}
\end{minipage}

\caption{\textbf{Fingerprint evaluation.} (a) Evaluation on \emph{Effectiveness}. True positive rate of model identification using fingerprints from four different pre-training LLMs (higher is better). Rates are computed across 10 fingerprints.
(b) Evaluation on \emph{Robustness}. True positive rate of model identification of finetuned models using fingerprints from four different pre-training LLMs (higher is better). Rates are computed across five different finetuned versions of the models and 10 fingerprints. See Table \ref{tab:downstream_rofl_variant} for studies on variant of \sys. See Appendix~\ref{appendix:detailed_results_rofl} for detailed results on each downstream finetuned versions of the models.}
\label{tab:TPR_combined}
\vspace{-2mm}
\end{table*}

\paragraph{Baselines.} 

\emph{(1.) Watermarking.} We include Instructional finetuning \textbf{(IF-SFT)}~\cite{xu-etal-2024-instructional}, which fully finetunes the model on a dataset containing fingerprint pairs, and \textbf{(IF-Emb)}, which restricts finetuning to the embedding layer~\cite{kurita2020weightpoisoningattackspretrained, yang-etal-2021-rethinking}. IF-SFT tends to overfit and shift parameters substantially, harming model performance, while IF-Emb mitigates this by limiting updates to embeddings. Both of these methods involve training the whole model with a fingerprint dataset.   
\emph{(2.) GCG-based fingerprinting.} We also consider the \textbf{GCG} method, which learns a random trigger string for a predefined target on a single model. We adopt the default learning setup by using 500 training epochs and stopping at the first success.
\emph{(3.) Variants of \sys.} Finally, we compare to simplified versions of our method: \textbf{Base-only}, which applies our target search and $n$-trial optimization to a single model; \textbf{Base + 1 task}, which learns fingerprints jointly on the base model and one finetuned task model; and \textbf{Base + 2 tasks}, which extends this to two finetuned models.  

\paragraph{Success measure.} Following~\citet{xu-etal-2024-instructional}, we measure the quality of model identification in terms of the true positive rate (TPR).
The TPR is defined as the rate of correct model identifications (that is, the true positive rate) across a set of $N$ fingerprints generated using that model.

\subsection{Experimentsal Results}

\noindent \textbf{$\bullet$ Effectiveness.} We evaluate four variants of GCG-based fingerprinting and two IF-based baselines in terms of effectiveness and persistence.
Table~\ref{tab:TPR_combined} (a.) reports the true positive rate (TPR) on four base models: Llama 2 7B, Llama 2 13B, Llama 3 8B, and Mistral 7B.
Our method consistently achieves 100\% TPR across all four models, outperforming other GCG baselines and remaining competitive with IF-SFT.
Compared with IF-Emb, which only updates embedding parameters, our method achieves stronger memorization without modifying model weights.
Unlike IF-SFT, which may degrade downstream utility, our method preserves standard performance while maintaining 100\% effectiveness.


\begin{table}[t]
\scriptsize
\setlength{\tabcolsep}{2pt} 
\renewcommand{\arraystretch}{1.2} 
\centering
\begin{tabular}{lccccc}
\hline
\textbf{Method}         & \textbf{\begin{tabular}[c]{@{}c@{}}Finetuned \\ Setting\end{tabular}} & \textbf{Llama 2 7B} & \textbf{Llama 2 13B} & \textbf{Llama 3 8B} & \textbf{Mistral 7B} \\ \hline
\textbf{Base}           & w/o                                                                   & 100\%              & 80\%                & 80\%               & 100\%               \\
\textbf{}               & SFT                                                                   & 94.58\%            & 83.33\%             & 80\%            & 96.67\%             \\
\textbf{}               & LoRA                                                                  & 100\%           & 90\%             & 83.33\%            & 100\%            \\ \hline
\textbf{+ 1 task}  & w/o                                                                   & 100\%           & 100\%            & 100\%           & 100\%            \\
                        & SFT                                                                   & 96.67\%           & 93.33\%             & 90\%            & 93.33\%             \\
\textbf{}               & LoRA                                                                  & 100\%           & 93.33\%             & 100\%           & 100\%            \\ \hline
\textbf{+ 2 tasks} & w/o                                                                   & 100\%           & 100\%            & 100\%           & 100\%            \\
                        & SFT                                                                   & 100\%           & 93.33\%             & 100\%           & 100\%            \\
\textbf{}               & LoRA                                                                  & 100\%           & 100\%            & 100\%           & 100\%            \\ \hline
\end{tabular}
\caption{Variant of \sys. We adopt multi-task learning on fingerprint optimization and verify various downstream finetuned models, including finetuning with SFT or LoRA. We report true positive rate of model identification using fingerprints from four different pre-training LLMs (higher is better). Rates are computed across 10 fingerprints.}
\label{tab:downstream_rofl_variant}
\vspace{-3mm}
\end{table}

\noindent \textbf{$\bullet$ Robustness.}
Table~\ref{tab:TPR_combined} (b.) compares robustness under downstream finetuning.
For each method, we average TPR over five downstream models derived from each base model using Alpaca-style finetuning settings.
Our method achieves 92$\sim$100\% TPR across all settings and consistently outperforms existing baselines.
In contrast, IF-SFT and IF-Emb suffer 10--35\% TPR degradation after finetuning.
Increasing the number of optimization tasks further improves robustness, demonstrating the effectiveness of multi-task fingerprint optimization.
Compared with standard GCG, our method improves TPR by 37$\sim$80\%, validating the importance of rare-region search and multi-trial optimization.
Furthermore, Table~\ref{tab:downstream_rofl_variant} analyzes the effect of multi-task optimization in \sys.
Increasing the number of optimization tasks from zero to two consistently improves TPR across all four models under both SFT and LoRA finetuning settings.
Additional variant results are provided in Appendix~\ref{appendix:detailed_results_rofl}.


\noindent \textbf{$\bullet$ Uniqueness.}
We further evaluate fingerprint verification in a fully black-box setting where suspect models are deployed as APIs or released on HuggingFace.
Table~\ref{tab:suspect_model_relevant} reports TPR on nine models finetuned from Llama 2 7B and FPR on five unrelated models.
For each suspect model, we use its repository-recommended prompt template and default decoding configuration.
Compared with GCG-QA~\cite{jin2024proflingofingerprintingbasedintellectualproperty} and the base-only variant, \sys successfully fingerprints all relevant models while rejecting unrelated ones (0\% FPR).

\begin{table}[t]
\centering
\scriptsize
\setlength{\tabcolsep}{3pt} 
\renewcommand{\arraystretch}{1.3} 
\begin{tabular}{lccc}
\hline
\textbf{\begin{tabular}[c]{@{}l@{}}Suspect Model /  Method\end{tabular}} & \multicolumn{1}{c}{\textbf{GCG-QA}} & \multicolumn{1}{c}{\textbf{Base}}  & \multicolumn{1}{c}{\textbf{\sys (Ours)}} \\ \hline
\cellcolor[HTML]{d5f5e3 }{\textbf{LLaMA-2-7b-chat-hf}}                                                  & \greencheck (20\%)                          &  \greencheck (40\%)                                                &  \greencheck (60\%)                              \\
\cellcolor[HTML]{d5f5e3 }{\textbf{LLaMA2-Chinese-7b-Chat}}                                            &  \greencheck(40\%)                          & \redcross (0\%)                                                    & \greencheck (80\%)                              \\
\cellcolor[HTML]{d5f5e3 }{\textbf{ELYZA-Japanese-LLaMA-2-7b}}                                     & \greencheck  (50\%)                          & \redcross (0\%)                                                        & \greencheck (100\%)                             \\
\cellcolor[HTML]{d5f5e3 }{\textbf{LLaMA-2-7b-ft-instruct-es}}                                        &\greencheck (30\%)                          & \redcross (0\%)                                                             & \greencheck (60\%)                              \\
\cellcolor[HTML]{d5f5e3 }{\textbf{Meditron-7b}}                                                     & \greencheck (60\%)                          & \redcross (0\%)                                               & \greencheck  (20\%)                              \\
\cellcolor[HTML]{d5f5e3 }{\textbf{Orca-2-7b}}                                                      & \greencheck  (30\%)                          & \redcross (0\%)                                               & \greencheck (40\%)                              \\
\cellcolor[HTML]{d5f5e3 }{\textbf{LLaMA-2-7b-DPO-v0.1}}                                               & \greencheck (30\%)                          & \greencheck (20\%)                                                       & \greencheck(40\%)                              \\
\cellcolor[HTML]{d5f5e3 }{\textbf{LLaMA-2-7b-DPO-Full-wo-Live-QA}}                                            & \greencheck(100\%)                          &\greencheck (80\%)                                                    & \greencheck (100\%)                             \\
\cellcolor[HTML]{d5f5e3 }{\textbf{TULU-2-DPO-7B}}                                                    & \greencheck(40\%)                          &\greencheck (20\%)                                      & \greencheck(40\%)                              \\ \hline
\cellcolor[HTML]{fadbd8}{\textbf{mistralai/mistral-7B-v0.3}}                                               &  \redcross (10\%)                          & \greencheck (0\%)                                                              & \greencheck(0\%)                                  \\
\cellcolor[HTML]{fadbd8}{\textbf{OpenAssistant/oasst-sft-4-pythia-12b}}                                                  &  \redcross (10\%)                           & \greencheck  (0\%)                                               & \greencheck (0\%)                                 \\
\cellcolor[HTML]{fadbd8}{\textbf{tiiuae/falcon-7b-instruct}}                                                 & \greencheck (0\%)                         &\greencheck (0\%)                                              & \greencheck (0\%)                               \\
\cellcolor[HTML]{fadbd8}{\textbf{TheBloke/guanaco-7B-HF}}                                                   & \greencheck (0\%)                       & \greencheck (0\%)                                                 & \greencheck (0\%)                             \\
\cellcolor[HTML]{fadbd8}{\textbf{mosaicml/mpt-7b-chat}}                                                  &  \greencheck (0\%)                        & \greencheck (0\%)                           & 
\greencheck (0\%)   \\

                         \hline
\end{tabular}
\caption{Evaluation on \emph{Uniqueness}. TPR on nine suspected \emph{relevant} (color with green margin) and FNR on five \emph{irrelevant} (color with red margin) models published in Hugging Face. For the baseline GCG-QA~\cite{jin2024proflingofingerprintingbasedintellectualproperty}, we evaluate on top of 10 QA pairs. 
For \emph{relevant} / \emph{irrelevant} models, higher TPR and lower FPR have better results. To show the uniqueness, we mark the results with the checkmark "\greencheck" if the relevant model is successfully fingerprinted. For irrelevant models, we mark the results with the checkmark "\greencheck" if they are 100\% not being fingerprinted (0\% on FPR).}
\label{tab:suspect_model_relevant}
\end{table}

\vspace{-2mm}
\section{Ablation Studies}
\label{sec:ablation}

\begin{table}[t]
\scriptsize
\setlength{\tabcolsep}{2pt} 
\renewcommand{\arraystretch}{1.4} 
\centering
\begin{tabular}{lcccc}
\hline
\textbf{Method}                       & \textbf{$\lambda$}      & \textbf{\begin{tabular}[c]{@{}c@{}}Filter Rejection  \\  Rate ($\tau=0.95$)\end{tabular}} & \multicolumn{2}{c}{\textbf{Llama 2 7B}}                    \\ \cline{4-5} 
                                      & \multicolumn{1}{l}{} & \multicolumn{1}{l}{}                                                   & \multicolumn{1}{l}{Base TPR} & \multicolumn{1}{l}{SFT TPR} \\ \hline
\textbf{RAFP w/o PPL}                         & 0                    & 100\%                                                                  & 100\%                        & 100\%                       \\ \hline
\multirow{3}{*}{\textbf{RAFP w/ PPL}} & \textbf{0.1}         & \textbf{20\%}                                                          & \textbf{90\%}                & \textbf{80\%}               \\
                                      & 0.3                  & 10\%                                                                   & 90\%                         & 60\%                        \\
                                      & 0.5                  & 10\%                                                                   & 80\%                         & 60\%                        \\ \hline
\end{tabular}
\caption{
Robustness of \sys\ under perplexity-based filtering. 
We compare the original \sys (\textit{w/o PPL}) with perplexity-aware optimization (\textit{w/ PPL}) under different regularization weights $\lambda$.
}
\label{tab:ppl_optimize}
\vspace{-4mm}
\end{table}

\paragraph{$\bullet$ Robustness against Perplexity Filtering.} To evaluate whether \sys can evade API-side perplexity filtering,
we also optimize fingerprints with a joint objective combining fingerprint consistency and prompt perplexity loss (PPL) regularization:
$
\mathcal L
=
-\log p_\theta(y\mid x)
+
\lambda\,\mathrm{PPL}(x)
$
, where $\lambda$ is the regularization weight. 
In our experiment, we set the filter threshold ($\tau$) for Llama 2 7B using the 95th percentile perplexity of 1K natural prompts sampled from ShareGPT dataset. This ensures that the filter accepts the realistic inputs while still detecting extremely low-likelihood prompts. The filter rule is $\text{reject}(x) = \text{PPL}(x) > \tau$.
In Table~\ref{tab:ppl_optimize},
we show \sys under perplexity-aware optimization substantially reduces rejection rates while retaining moderate fingerprint effectiveness on both the base and SFT-adapted models. We grid search $\lambda$ from 0.1 to 0.5 and observe that $\lambda=0.1$  lowers the reject rate without severely degrading TPR.

\begin{table}[t]
\scriptsize
\setlength{\tabcolsep}{2pt} 
\renewcommand{\arraystretch}{1.4} 
\centering
\begin{tabular}{lc|ccccc}
\hline
\textbf{Model}                                                                        & \textbf{\begin{tabular}[c]{@{}c@{}}Prompt / \\ Method\end{tabular}} & \textbf{Llama 2} & \textbf{Vicuna} & \textbf{Alpaca} & \textbf{Zero Shot} & \textbf{ChatGPT} \\ \hline
\multirow{3}{*}{\textbf{\begin{tabular}[c]{@{}l@{}}Llama 2 7B \\ (Base)\end{tabular}}} & \textbf{IF-SFT}                                                              & 62.5\%           & \textbf{100\%}  & 75\%            & 75\%                & 50\%             \\    & \textbf{GCG}                                                                 & \textbf{100\%}   & 80\%            & \textbf{100\%}  & 60\%                & 20\%             \\
 & \textbf{\sys}                                                                & \textbf{100\%}   & \textbf{100\%}            & 90\%  & \textbf{80\%}       & \textbf{100\%}   \\ \hline
\multirow{3}{*}{\textbf{\begin{tabular}[c]{@{}l@{}}Finetuned \\ (ShareGPT)\end{tabular}}}                                                   & \textbf{IF-SFT}                                                              & 75\%             & \textbf{100\%}  & \textbf{100\%}  & 62.5\%              & 50\%             \\
                                                                                      & \textbf{GCG}                                                                 & \textbf{100\%}   & 50\%            & 90\%            & 60\%                & 10\%             \\
                                                                                      & \textbf{\sys}                                                                & \textbf{100\%}   & \textbf{100\%}  & 90\%            & \textbf{70\%}       & \textbf{100\%}   \\ \hline
\end{tabular}
\vspace{-2mm}
\caption{Analysis of different prompt templates: We demonstrate the robustness of \sys with respect to changes in prompt templates. Compared to two other baselines, \sys shows the smallest drop in TPR.}
\label{tab:prompt_analysis}
\end{table}

\begin{figure}[t]
    \centering
\includegraphics[width=0.99\linewidth]{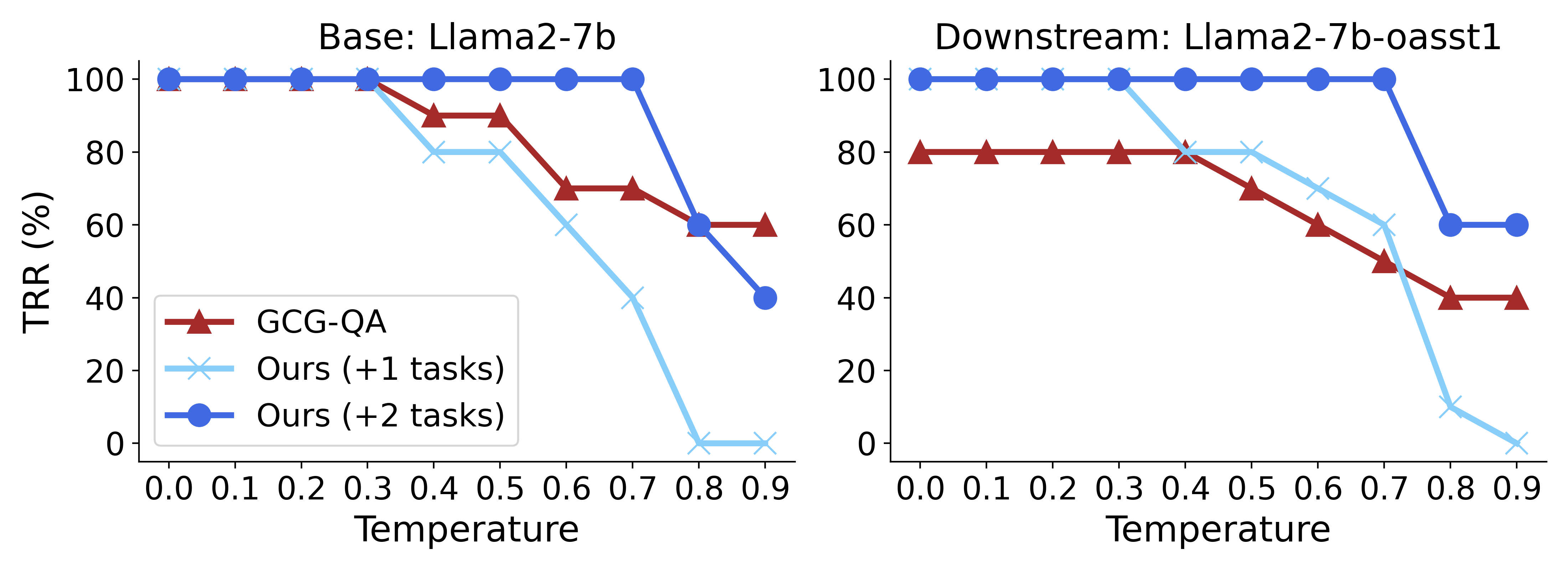}
       \caption{Analysis of generation hyperparameters: We set up temperatures for the model during inference time to demonstrate the robustness of \sys under various temperatures. We discovered that \sys (+ 2 tasks) maintains a high TPR, which slightly drops after the temperature reaches 0.8.}
    \label{fig:temp_analysis}
\end{figure}
\vspace{-3mm}

\paragraph{$\bullet$ Analysis of Prompt Templates.} We evaluate the robustness of \sys\ under different prompt templates on Llama 2 7B and its downstream models.
Fingerprints are optimized using the default Llama 2 and Vicuna templates, and evaluated on additional templates including Alpaca, Zero-Shot, and ChatGPT~\cite{zheng2023judging}.
As shown in Table~\ref{tab:prompt_analysis}, \sys\ maintains 70--100\% TPR across both base and finetuned models under prompt-template variations.
Compared with GCG and IF-SFT, our method shows the smallest TPR degradation (10--30\%), demonstrating stronger robustness to prompt-format changes.
Detailed prompt templates are provided in Table~\ref{tab:promptdetails}.

\paragraph{$\bullet$ Analysis of Generation Hyperparameters.}
We evaluate robustness under different temperatures ranging from 0 to 0.9.
In Fig.~\ref{fig:temp_analysis}, \sys\ maintains nearly 100\% TPR from temperature 0.1 to 0.7, with only slight degradation above 0.8.
In contrast, GCG-QA begins to degrade at a temperature of 0.3.
Additional analyses on multi-stage fingerprints, quantization variants, and dissimilarity of bottom-k tokens are provided in~\ref{appendix:multi_stage_analyasis},~\ref{appendix:quantization_analysis}, and~\ref{supp:dis-bottomk}.

\vspace{-2mm}

\section{Conclusion}
\label{sec:conclusion}


In this work, we propose \sys, a method for identifying LLM lineages via rare-region fingerprints.
Our key insight is that rare prompt regions receive weak optimization signal during downstream finetuning, allowing rare prompt-response behaviors to remain stable across model adaptations.
We further provide a theoretical analysis explaining the invariance of rare-region fingerprints under finetuning.
Experiments across multiple LLM families and downstream adaptations demonstrate that RAFP achieves strong fingerprint persistence and substantially outperforms prior fingerprinting baselines in black-box settings.
Future work may explore stronger adaptive adversaries and fingerprint LLMs with web-scale model training.

\section{Limitation}
\label{limitation}

\textbf{$\bullet$ Vulnerability to Strong Adaptations.} Fingerprinting exploits statistical patterns that survive common changes (e.g., finetuning, quantization), but stronger adaptations such as distillation, re-parameterization, or adversarial finetuning can weaken or erase fingerprints. These attacks are costly but remain a potential threat.
Besides, an adversary may plant data in the public domain that enters training, then commits to the resulting fingerprints first (cf. data poisoning). Preliminary discussion is in Appendix~\ref{supp:frontrunning}.  


\noindent \textbf{$\bullet$ Black-Box Constraints and Practicality.}
Fingerprinting in a black-box setting requires querying the suspect model, which faces two challenges: limited or costly API access, and filtering or post-processing that can suppress fingerprints without degrading performance (see Appendix~\ref{supp:filteringattack}).

\section{Ethical Consideration}
\label{ethicalconsider}


Technically, \sys serves as a robust method to identify and track the origin of specific models involved in generating particular texts. This capability enhances models' transparency, aids in attributing content to its source, and can deter the misuse of LLMs. 

Ethically, fingerprinting LLM might raise concerns about surveillance and privacy. It requires carefully weighing the need for security and accountability against the risk of violating users' privacy rights.
By allowing the identification of downstream models' origins, there is a risk that this capability could be exploited for surveillance and monitoring, potentially resulting in limiting the right of users on using public models or the suppression of free speech.

Future societal consequences of deploying fingerprints with LLMs hinge on regulatory and governance frameworks. People need to have clear policies that define acceptable uses of LLMs and further recognize which form of fingerprinting can be used for model verification. We believe this technology can foster a more trustworthy digital communication environment, reducing the prevalence of fake news and enhancing the credibility of LLM-generated content.

\bibliography{custom}

\clearpage

\appendix

\section{Appendix}

\subsection{Detailed Results of \sys on Downstream Tuning (DT)} 
\label{appendix:detailed_results_rofl}
In the main paper, Table~\ref{tab:TPR_combined}, we show the average scores for different models on downstream finetuning. Here, in Table~\ref{tab:detailed_baseonly},\ref{tab:detailed_baseplusone}, and~\ref{tab:detailed_baseplustwo}, we show the detailed results of \sys under three different settings: \textbf{base only}, \textbf{base + 1 task}, and \textbf{base + 2 tasks}. For each table, we show the performance on five downstream models, including sharegpt, ni, dolly, codegen, and oasst1. In each table, we show variants of tuning methods, including SFT and LoRA. For variants of tuning methods, we set the learning rate for SFT as 2e-5 and the training epoch as 3. For LoRA, we set the alpha as 32 and r as 16.
\paragraph{The details of downstream finetuned tasks}
\begin{itemize}

  \item \textbf{ShareGPT}: The dataset consists of approximately 70K user-shared conversations, including around 6000 expert conversations generated by GPT-4 and the sub-optimal conversations from GPT-3.5.
  
    \item \textbf{ni}: \texttt{Natural Instruction} is a comprehensive dataset encompassing 61 distinct tasks, featuring human-authored instructions for each and totaling 193,000 task instances. These instructions were sourced from crowdsourcing, originally used to develop existing NLP datasets, and have been systematically organized into a unified schema to facilitate diverse NLP applications.
    
    \item \textbf{dolly}: \texttt{databricks-dolly-15k} is an open-source dataset created by Databricks, featuring thousands of instruction-following records generated by its employees, covering several behavioral categories like brainstorming, classification, QA, generation, information extraction, and summarization as outlined in the InstructGPT paper.

    
    \item \textbf{codegen}: \texttt{
evol-codealpaca-v1} dataset contains 20,000 instruction-following entries specifically used to finetune the Code Alpaca model. This project aims to develop and distribute an instruction-following LLaMA model that specializes in code generation, enhancing the capabilities of AI in software development contexts.
    \item \textbf{oasst1}: \texttt{OpenAssistant Conversations (OASST1)} is a human-generated and human-annotated dataset comprising 161,443 messages across 35 different languages. It includes 461,292 quality ratings and results in over 10,000 fully annotated conversation trees, providing a rich resource for developing and refining multilingual assistant-style conversational models.
\end{itemize}
\begin{table}[t]
\scriptsize
\setlength{\tabcolsep}{3pt} 
\renewcommand{\arraystretch}{1.4} 
\centering
\begin{tabular}{llcccc}
\hline
\textbf{\sys (Base)}                                                                      &               & \multicolumn{4}{c}{\textbf{Model}}                                                     \\
                                                                                               &               & \textbf{Llama 2 7B} & \textbf{Llama 2 13B} & \textbf{Llama 3 8B} & \textbf{Mistral 7B} \\ \hline
\textbf{TRR (pre)}                                                                             &               & \textbf{100.00\%}   & \textbf{80.00\%}     & \textbf{80.00\%}    & \textbf{100.00\%}   \\ \hline
\multirow{7}{*}{\textbf{\begin{tabular}[c]{@{}l@{}}TRR (post)\\ DT-SFT\end{tabular}}}  & sharegpt      & 87.50\%             & 60.00\%              & 60.00\%             & 100.00\%            \\
                                                                                               & ni            & 100.00\%            & 80.00\%              & 80.00\%             & 100.00\%            \\
                                                                                               & dolly         & 100.00\%            & 100.00\%             & 80.00\%             & 100.00\%            \\
                                                                                               & codegen       & 80.00\%            & 80.00\%             & 60.00\%             & 100.00\%                \\
                                                                                               & oasst1        & 100.00\%            & 100.00\%             & 100.00\%            & 100.00\%            \\
                                                                                               & \textbf{avg.} & \textbf{94.58\%}    & \textbf{83.33\%}     & \textbf{80.00\%}    & \textbf{96.67\%}    \\ \hline
\multirow{7}{*}{\textbf{\begin{tabular}[c]{@{}l@{}}TRR (post)\\ DT-LoRA\end{tabular}}} & sharegpt      & 100.00\%            & 80.00\%              & 60.00\%             & 100.00\%            \\
                                                                                               & ni            & 100.00\%            & 100.00\%             & 80.00\%             & 100.00\%            \\
                                                                                               & dolly         & 100.00\%            & 100.00\%             & 100.00\%            & 100.00\%            \\
                                                                                               & codegen       & 100.00\%            & 80.00\%              & 60.00\%             & 100.00\%            \\
                                                                                               & oasst1        & 100.00\%            & 100.00\%             & 100\%               & 100.00\%            \\
                                                                                               & \textbf{avg.} & \textbf{100.00\%}   & \textbf{90.00\%}     & \textbf{83.33\%}    & \textbf{100.00\%}   \\ \hline
\end{tabular}
\caption{Detailed results for Base only}
\label{tab:detailed_baseonly}
\end{table}

\begin{table}[t]
\setlength{\tabcolsep}{3pt} 
\renewcommand{\arraystretch}{1.4} 
\scriptsize
\centering
\begin{tabular}{llrrrr}
\hline
\textbf{\sys (+ 1 task)}                                                                  &               & \multicolumn{4}{c}{\textbf{Model}}                                                                                                                                     \\
                                                                                               &               & \multicolumn{1}{c}{\textbf{Llama 2 7B}} & \multicolumn{1}{c}{\textbf{Llama 2 13B}} & \multicolumn{1}{c}{\textbf{Llama 3 8B}} & \multicolumn{1}{c}{\textbf{Mistral 7B}} \\ \hline
\textbf{TRR (pre)}                                                                             &               & \textbf{100.00\%}                       & \textbf{100.00\%}                        & \textbf{100.00\%}                       & \textbf{100.00\%}                       \\ \hline
\multirow{7}{*}{\textbf{\begin{tabular}[c]{@{}l@{}}TRR (post)\\ DT-SFT\end{tabular}}}  & sharegpt      & 100.00\%                                & 100.00\%                                 & 90.00\%                                 & 100.00\%                                \\
                                                                                               & ni            & 100.00\%                                & 80.00\%                                  & 80.00\%                                 & 60.00\%                                 \\
                                                                                               & dolly         & 100.00\%                                & 80.00\%                                  & 90.00\%                                 & 100.00\%                                \\
                                                                                               & codegen       & 80.00\%                                & 100.00\%                                  & 80.00\%                                & 100.00\%                                \\
                                                                                               & oasst1        & 100.00\%                                & 100.00\%                                 & 100.00\%                                & 100.00\%                                \\
                                                                                               & \textbf{avg.} & \textbf{96.67\%}                       & \textbf{93.33\%}                         & \textbf{90.00\%}                        & \textbf{93.33\%}                        \\ \hline
\multirow{7}{*}{\textbf{\begin{tabular}[c]{@{}l@{}}TRR (post)\\ DT-LoRA\end{tabular}}} & sharegpt      & 100.00\%                                & 100.00\%                                 & 100.00\%                                & 100.00\%                                \\
                                                                                               & ni            & 100.00\%                                & 100.00\%                                 & 100.00\%                                & 100.00\%                                \\
                                                                                               & dolly         & 100.00\%                                & 80.00\%                                  & 100.00\%                                & 100.00\%                                \\
                                                                                               & codegen       & 100.00\%                                & 100.00\%                                  & 100.00\%                                & 100.00\%                                \\
                                                                                               & oasst1        & 100.00\%                                & 100.00\%                                 & 100.00\%                                & 100.00\%                                \\
                                                                                               & \textbf{avg.} & \textbf{100.00\%}                       & \textbf{93.33\%}                         & \textbf{100.00\%}                       & \textbf{100.00\%}                       \\ \hline
\end{tabular}
\caption{Detailed results for Base + 1 task.}
\label{tab:detailed_baseplusone}
\end{table}

\begin{table*}[h]
\centering
\small
\setlength{\tabcolsep}{3pt} 
\renewcommand{\arraystretch}{1.4} 
\begin{tabular}{ll}
\hline
\textbf{Prompt}                              & \textbf{Message}                                                                                                                                                                                                                                                     \\ \hline
\textbf{Llama 2}                             & \begin{tabular}[c]{@{}l@{}}\textless{}s\textgreater{}{[}INST{]}\\ {[}/INST{]}\end{tabular}                                                                                                                                                                        \\ \hline

\textbf{Vicuna\_v1.1} & \begin{tabular}[c]{@{}l@{}}\textless{}s\textgreater A chat between a curious user and an artificial intelligence assistant. \\ The assistant gives helpful, detailed, and polite answers to the user questions.\end{tabular}                \\ \hline
\textbf{Alpaca}                              & \begin{tabular}[c]{@{}l@{}}\textless{}s\textgreater Below is and instruction that describes a task. Write a response that appropriately completes the request \\ \#\#\# Instruction:\\ \#\#\# Response:\end{tabular}                                                 \\ \hline
\textbf{Zero-Shot}                           & \begin{tabular}[c]{@{}l@{}}\textless{}s\textgreater A chat between a curious human and an artificial intelligence assistant. \\ The assistant gives helpful, detailed, and polite answers to the human's questions.\\ \#\#\# Human:\\ \#\#\# Assistant:\end{tabular} \\ \hline
\textbf{ChatGPT}                             & \begin{tabular}[c]{@{}l@{}}\textless{}s\textgreater You are a helpful assistant\\ user:\\ assistant:\end{tabular}                                                                                                                                                    \\ \hline
\end{tabular}
\caption{Details of different system prompt messages.}
\label{tab:promptdetails}
\end{table*}

\subsection{Details of modified prompts}
In the main paper, Table~\ref{tab:prompt_analysis} shows that \sys is robust to changes in different system prompt templates. Here, in Table~\ref{tab:promptdetails}, we demonstrate how different the modified prompt templates are.

\subsection{Robustness Analysis on Multi-Stage Fingerprint }
\label{appendix:multi_stage_analyasis}
We consider a more challenging scenario, where the users can finetune the model in multiple rounds using several downstream datasets. In this case, the fingerprints are highly possible being removed due to the large parameter shifts after finetuning. In Table~\ref{tab:multistage_result}, we sequentially finetune Llama 2 7B with three downstream tasks, and evaluate the model on each three stage. For our multi-learning methods, we learn the fingerprint on one or two of other subtasks, which do not overlap with the three tasks we use to evaluate. As the results show, after three stages of finetuning, both of our methods (+ 1 task and + 2 tasks) maintain 100\% TPR on base model and finetuned models. 

\begin{table}[t]
\centering
\scriptsize
\setlength{\tabcolsep}{2pt} 
\renewcommand{\arraystretch}{1.4} 
\begin{tabular}{lcc|llll}
\hline 
 \textbf{Model}                     & \multicolumn{2}{c|}{\textbf{Finetune Setting}}      & \multicolumn{4}{c}{\textbf{Method}}                                                                                                                                 \\
\multicolumn{1}{l}{}                                                                      & \textbf{Stage} & \multicolumn{1}{c|}{\textbf{Task}} & \multicolumn{1}{c}{\textbf{GCG}} & \multicolumn{1}{c}{\textbf{Base only}} & \multicolumn{1}{c}{\textbf{+ 1 task}} & \multicolumn{1}{c}{\textbf{+ 2 tasks}} \\ \hline
\textbf{\begin{tabular}[c]{@{}l@{}}LLaMA 2 7B\\ (base)\end{tabular}}                       & 0              & \multicolumn{1}{c|}{--}           & 25\%                                      & 100\%                                  & 100\%                                 & \textbf{100\%}                         \\ \hline
\multirow{3}{*}{\textbf{Finetuned}} & 1              & w/ ShareGPT                        & 20\%                                      & 88\%                                   & 100\%                                 & \textbf{100\%}                         \\
                                                                                          & 2              & w/ NI                              & 0\%                                       & 75\%                                   & 100\%                                 & \textbf{100\%}                         \\
                                                                                          & 3              & w/ Dolly                           & 0\%                                       & 62.50\%                                & 100\%                                 & \textbf{100\%}                         \\ \hline
\end{tabular}
\caption{Analysis on the effectiveness and robustness for multi-stage fingerprinting}
\label{tab:multistage_result}
\end{table}

\subsection{Robustness Analysis on Variant of Quantization}
\label{appendix:quantization_analysis}
\begin{figure}[t]
\centering
\includegraphics[width=0.50\textwidth]{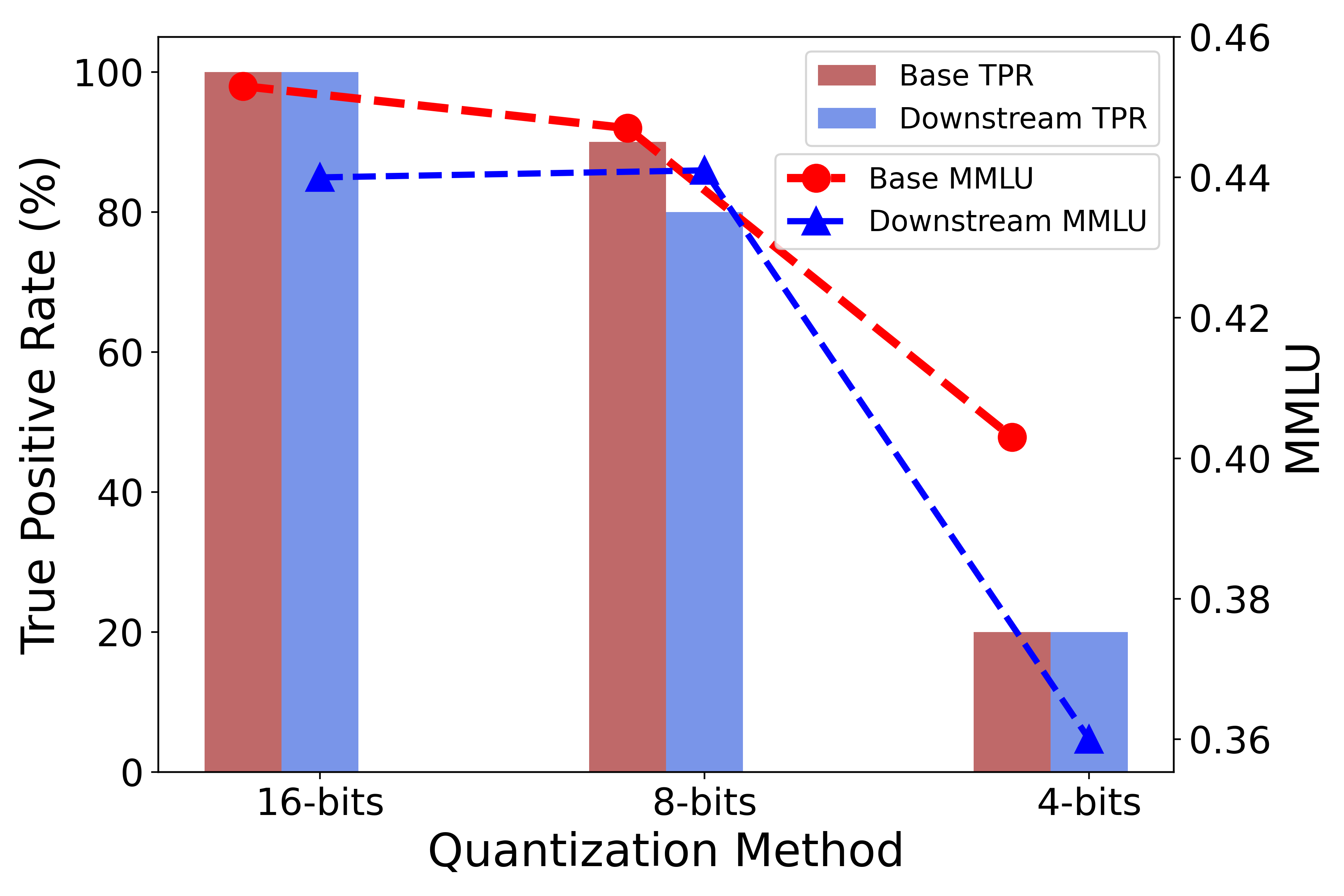}
\caption{Variant of quantization. Evaluate on base/downstream tuning model under different quantization setting}
\vspace{-1ex}
\label{fig:quantized_Analysis}
\end{figure}

We further investigate more advanced adaptation methods, such as quantization (commonly used to improve inference efficiency. We evaluate the robustness of \sys on different quantization settings, starting from \textit{16-bit} to \textit{4-bit} on both base and downstream tuning models. In Fig.~\ref{fig:quantized_Analysis}, we demonstrate two metrics, including TPR and MMLU score, to show the trade-off between generation quality and fingerprint robustness. We adopt QLoRA with \textit{16-bit}, \textit{8-bits}. and \textit{4-bit} for quantization. As Fig.~\ref{fig:quantized_Analysis} shows, \sys maintains high TPR, and the MMLU score drops slightly when under \textit{16-bit} and \textit{8-bits} quantization. For \textit{4-bit} quantization, although the TPR drops significantly, the MMLU score drops correspondingly, severely decreasing the economic value of the model.

\begin{table}[t]
\centering
\scriptsize
\setlength{\tabcolsep}{2pt} 
\renewcommand{\arraystretch}{1.4} 
\begin{tabular}{llrrrr}
\hline
\textbf{\sys (+ 2 tasks)}                                                                 &               & \multicolumn{4}{c}{\textbf{Model}}                                                                                                                                     \\
                                                                                               &               & \multicolumn{1}{c}{\textbf{Llama 2 7B}} & \multicolumn{1}{c}{\textbf{Llama 2 13B}} & \multicolumn{1}{c}{\textbf{Llama 3 8B}} & \multicolumn{1}{c}{\textbf{Mistral 7B}} \\ \hline
\textbf{TRR (pre)}                                                                             &               & \textbf{100.00\%}                       & \textbf{100.00\%}                        & \textbf{100.00\%}                       & \textbf{100.00\%}                       \\ \hline
\multirow{7}{*}{\textbf{\begin{tabular}[c]{@{}l@{}}TRR (post)\\ DT-SFT\end{tabular}}}  & sharegpt      & 100.00\%                                & 80.00\%                                  & 100.00\%                                & 100.00\%                                \\
                                                                                               & ni            & 100.00\%                                & 100.00\%                                 & 100.00\%                                & 100.00\%                                \\
                                                                                               & dolly         & 100.00\%                                & 100.00\%                                 & 100.00\%                                & 100.00\%                                \\
                                                                                               & codegen       & 100.00\%                                & 100.00\%                                 & 100.00\%                                & 100.00\%                                \\
                                                                                               & oasst1        & 100.00\%                                & 80.00\%                                  & 100.00\%                                & 100.00\%                                \\
                                                                                               & \textbf{avg.} & \textbf{100.00\%}                       & \textbf{92.00\%}                         & \textbf{100.00\%}                       & \textbf{100.00\%}                       \\ \hline
\multirow{7}{*}{\textbf{\begin{tabular}[c]{@{}l@{}}TRR (post)\\ DT-LoRA\end{tabular}}} & sharegpt      & 100.00\%                                & 100.00\%                                 & 100.00\%                                & 100.00\%                                \\
                                                                                               & ni            & 100.00\%                                & 100.00\%                                 & 100.00\%                                & 100.00\%                                \\
                                                                                               & dolly         & 100.00\%                                & 100.00\%                                 & 100.00\%                                & 100.00\%                                \\
                                                                                               & codegen       & 100.00\%                                & 100.00\%                                 & 100.00\%                                & 100.00\%                                \\
                                                                                               & oasst1        & 100.00\%                                & 100.00\%                                 & 100.00\%                                & 100.00\%                                \\
                                                                                               & \textbf{avg.} & \textbf{100.00\%}                       & \textbf{100.00\%}                        & \textbf{100.00\%}                       & \textbf{100.00\%}                       \\ \hline
\end{tabular}
\caption{Detailed results for Base + 2 tasks}
\label{tab:detailed_baseplustwo}
\end{table}

\subsection{Dis-similarlity of bottom-k tokens}
\label{supp:dis-bottomk}
To verify how much dis-similarlity for the set of bottom-k tokens for given models if the supported languages are held constant, we collect the last 2k tokens from each base model and calculate the Jaccard Similarity between them. The Jaccard Similarity is defined as the size of the intersection divided by the size of the union of two sets. As the Table~\ref{tab:dissim_metric} shows, the set similarity between different model lineages is extremely small. Besides, in Appendix~\ref{appendix:unforge}, we calculate the possibility of generating the same fingerprint pairs with random query-response pairs. The probability of generating the same fingerprint y is $1/D^{|y|}$, where D is the token space of $2^k$ and $|y|$ is the token length, yielding a probability of $1.95e-30$.

\begin{table}[]
\centering
\setlength{\tabcolsep}{2pt} 
\renewcommand{\arraystretch}{1.7} 
\small
\begin{tabular}{l|ccc}
                   & \textbf{Llama 2 7B} & \textbf{Llama 3 8B} & \textbf{Mistral 7B} \\ \hline
\textbf{Llama 2 7B} & 1                  & 0.0021             & 0.4312           \\
\textbf{Llama 3 8B} & -                  & 1                  & 0.0024           \\
\textbf{Mistral 7B}   & -                  & -                  & 1               
\end{tabular}
\caption{Jaccard Similarity for the set of last 2k bottom tokens from three base models.}
\label{tab:dissim_metric}
\end{table}

\subsection{Front-running Attack Analysis}

\label{supp:frontrunning}


In our threat model, we assumed the adversary does not have any influence over the model development process. In this section, we relax this assumption and consider a \emph{front-running attack} based on data poisoning. Since LLMs are typically trained on web-crawled data, the attacker may inject fingerprints into the model in the following manner: 1) The attacker constructs a trigger-fingerprint pair and posts it on a public web domain. 2) The attacker commits this trigger-fingerprint pair according to the procedure described in Sec.\ref{sec:method}. 3) The LLM trainer crawls the web and constructs a training dataset that contains the trigger-fingerprint pair, and trains the model on this dataset. 4) When the LLM is released, the attacker claims ownership over the model by verifying the committed fingerprint according to Sec.\ref{sec:method}. This fingerprint has precedence over any fingerprint constructed by our method post-training, and thus the attacker is established as the model owner.

The front-running attack exploits the known data poisoning vulnerability of web-scale model training~\citep{carlini2024poisoning}, and can be plausibly executed against our fingerprinting protocol. 
To understand the threat of this front-running attack, we ask how much training data does attacker need to control to successfully execute the attack?
We simulate such an attack by finetuning the LLM on a fixed trigger-fingerprint pair until the LLM achieves 100\% TPR on the injected fingerprint. Since frontier LLMs often make only a single pass over its training data, the number of steps required to achieve 100\% TPR can be thought of as a lower bound for the number of training samples the attacker needs to control. Figure~\ref{fig:FrontRunningAttack} shows the number of steps required as a function of the length of the fingerprint. As the length of fingerprint increases, the attacker needs to control a larger number of training samples, although this number is relatively low even for fingerprints of length 54. Our experiment suggests that to prevent this front-running attack, the model trainer should select longer fingerprints and apply strong data de-duplication to reduce the attacker's control over training data.
\begin{figure}[t]

\centering
\includegraphics[width=0.40\textwidth]{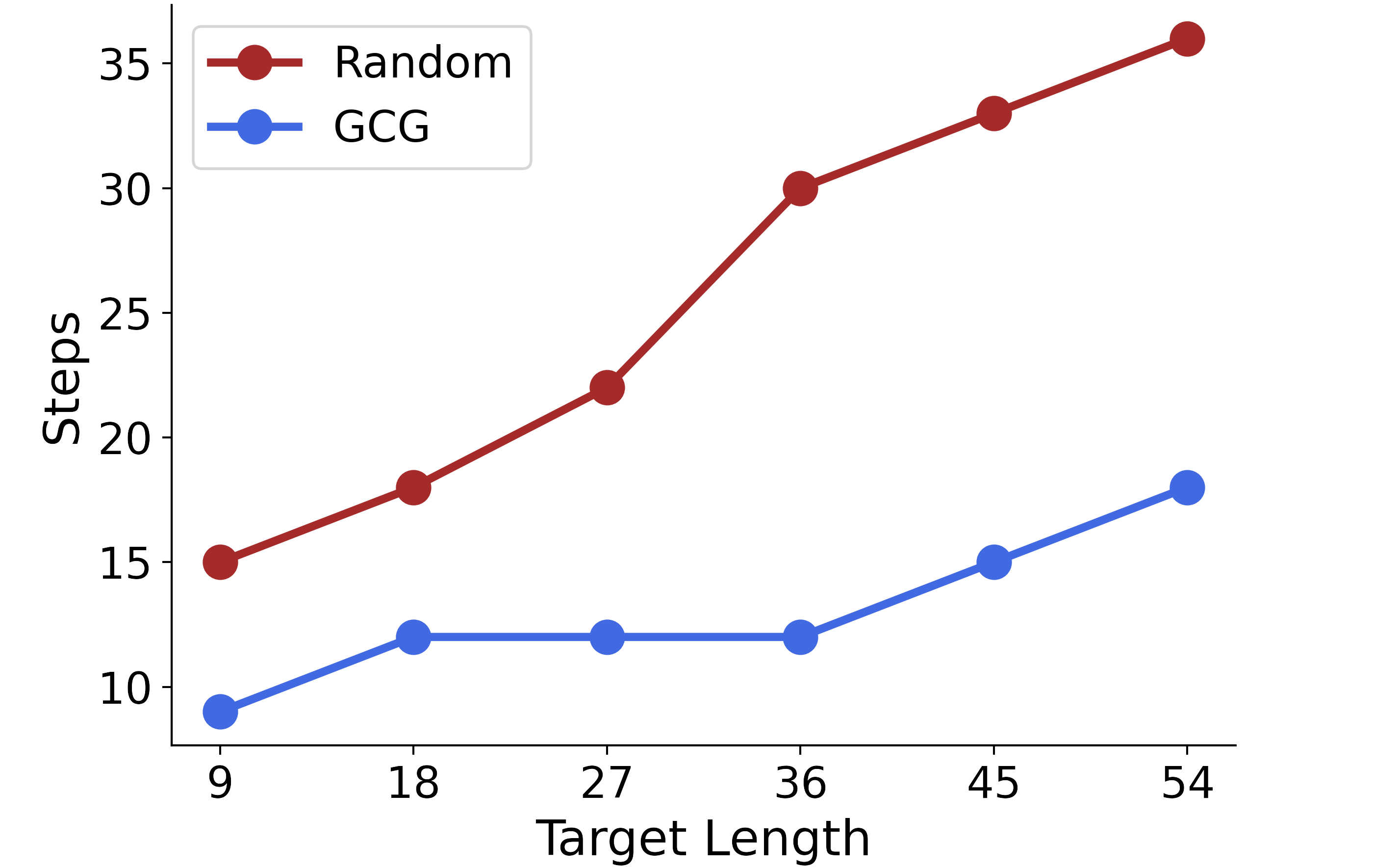}
\caption{Analysis of the front-running attack. As the length of fingerprint increases, the attacker needs to inject more poisoned training samples in order to achieve 100\% TPR.}

\label{fig:FrontRunningAttack}
\end{figure}

\begin{figure}[t]

\centering
\includegraphics[width=0.49\textwidth]{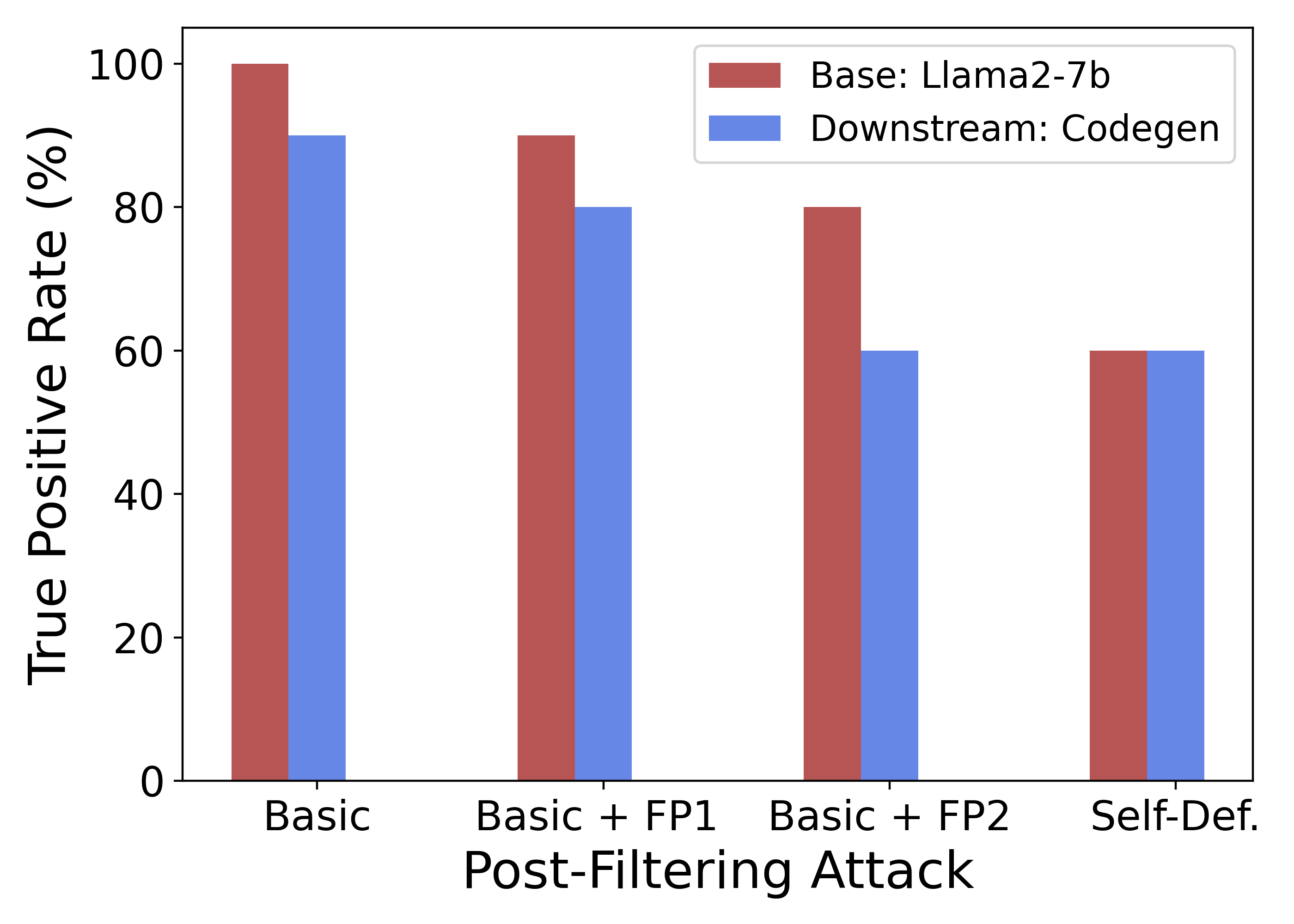}
\caption{Analysis of the filtering attack. The \textit{Basic} means we only use the basic prompt. The \textit{Basic + FP1} means we append the filter prompt 1 with the basic system prompt, and the \textit{Basic + FP2} means we append the filter prompt 2 with the basic system prompt. Our experiment shows that \sys maintains high TPR after post-filtering attack.}

\label{fig:FrontRunningAttack}
\end{figure}


\subsection{Pre- and Post-Filtering Attack Analysis}

\label{supp:filteringattack}

We discuss more potential risks such as pre-filtering and post-filtering. An attacker may want to pre-define a prompt to filter out the non-sensical response or leverage perplexity filtering to detect unreadable strings. The following shows the analysis under these scenarios.

\paragraph{Post-filtering with system prompts} Similar to the output filtering for jailbreak attacks on LLM~\cite{mo2024fightjailbreakingpromptadversarial}, we pre-define several system prompts involving filtering messages. We describe the prompts as follows. The placeholder in the format of the input message can be replaced by the messages mentioned above. We append the filter prompts with the basic prompt and concatenate with the fingeprint.

\vspace{2mm}
\begin{tcolorbox}[colback=white!5!white,colframe=gray!2!gray,arc=2mm]
\begin{adjustbox}{minipage=[c][1.60\linewidth]{1\linewidth}}
- Format of Input Message:  \{{\color{blue}\textbf{\texttt{Basic Prompt}}}\}, \{{\color{ForestGreen}\textbf{\texttt{Filter Prompt}}}\}, \textbf{\texttt{\#\#\# Instruction:}} \{{\color{BrickRed} \textbf{\texttt{Fingerprint}}}\} \textbf{\texttt{\#\#\# Response:}}
 \\ \par
- {\color{blue}\textbf{\texttt{Basic Prompt}}}: Below is an instruction that describes a task. Write a response that appropriately completes the request. \\ \par
- {\color{ForestGreen}\textbf{\texttt{Filter prompt 1}}}: You are a helpful, respectful, and honest assistant. Always answer as helpfully as possible, while being sensical. 
\\ \par
- {\color{ForestGreen}\textbf{\texttt{Filter prompt 2}}}: You are an assistant that is harmless and helpful. Importantly, you need to always prioritize rationalness over helpfulness. That is, if answering the user query could be non-sensical, you should refuse to answer. Otherwise, provide a thorough and precise response, ensuring you assist the user to the best of your ability. \\ \par

\end{adjustbox}
\end{tcolorbox}

\begin{figure}[t]

\centering
\includegraphics[width=0.49\textwidth]{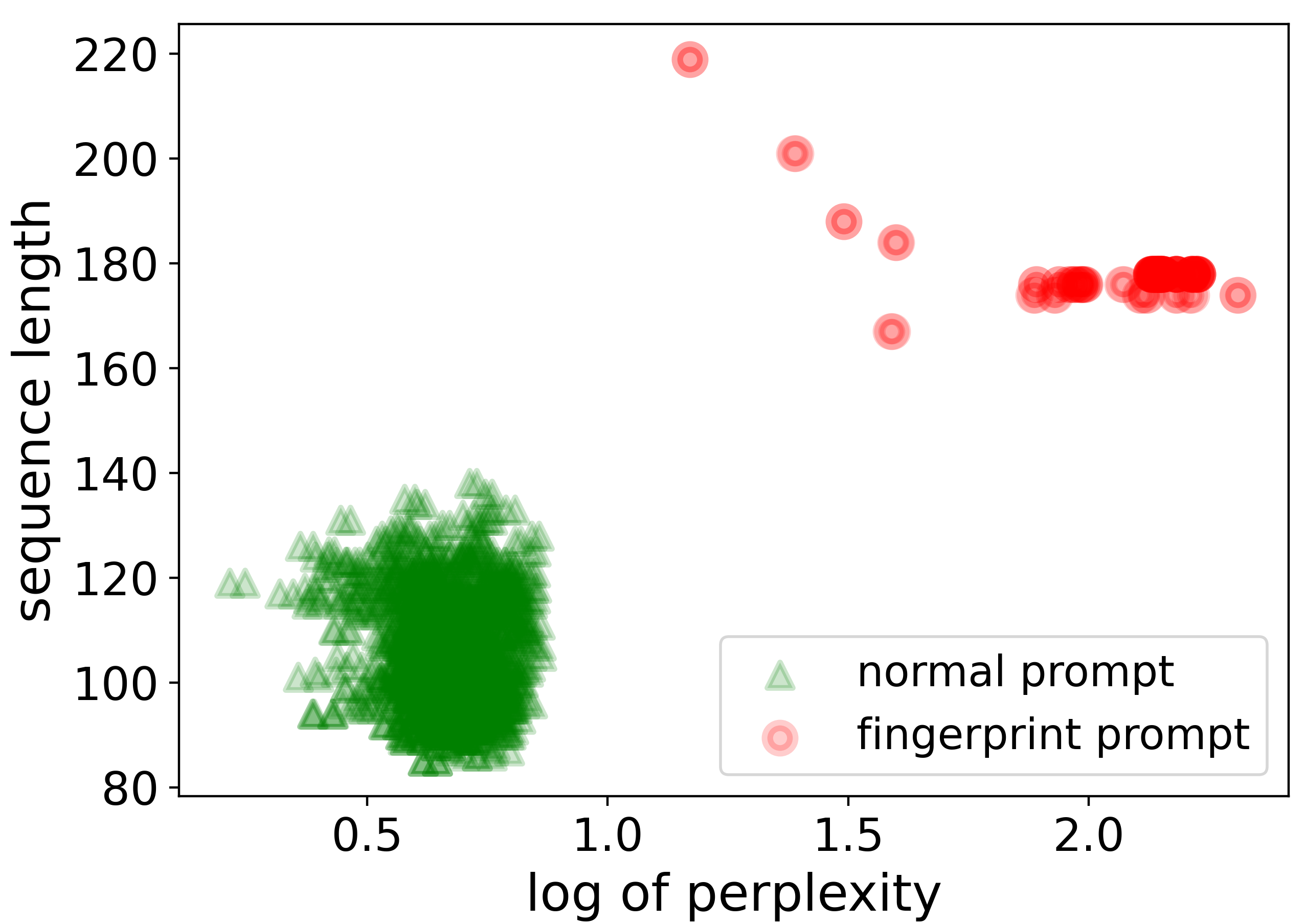}
\caption{Analysis of the perplexity filtering. We provide the perplexity analysis on the fingerprint input. The scatter plot shows a significant difference between two different prompts, which means it is possible to use perplexity to remove our fingerprint.}

\label{fig:FrontRunningAttack}
\end{figure}

\paragraph{Self-Defend~\cite{caoetal2024defending}}: The potentially non-sensical response generated from an LLM is inserted into a prompt format and provided to a filter. The filter is given instructions to classify the response as either "normal" or "abnormal". The filter message for self-defend is defined as \textit{"Is this an unlikely or non-sensical response of an LLM?"}

\paragraph{Perplexity (PPL) filtering} By evaluating the perplexity of queries with unreadable strings using an open-source LLM (GPT-2), the literature observes that these strings have exceedingly high perplexity values~\cite{alon2023detectinglanguagemodelattacks}. Motivated by this, we provide the perplexity 
on the fingerprint input.
For the normal prompt, we collect 900 normal input prompts from downstream subtask CodeGen and compute the corresponding perplexity. For our fingerprint promps, we calculate perplexity over all the fingerprint we learned from Llama 2 7B. We plot the PPL v.s. length of sequence for each prompt. The scatter plot shows a significant difference between two different prompts, which means it is possible to use perplexity to remove our fingerprint.


\begin{figure*}[t]
\centering
\centering
\includegraphics[width=0.99\textwidth]{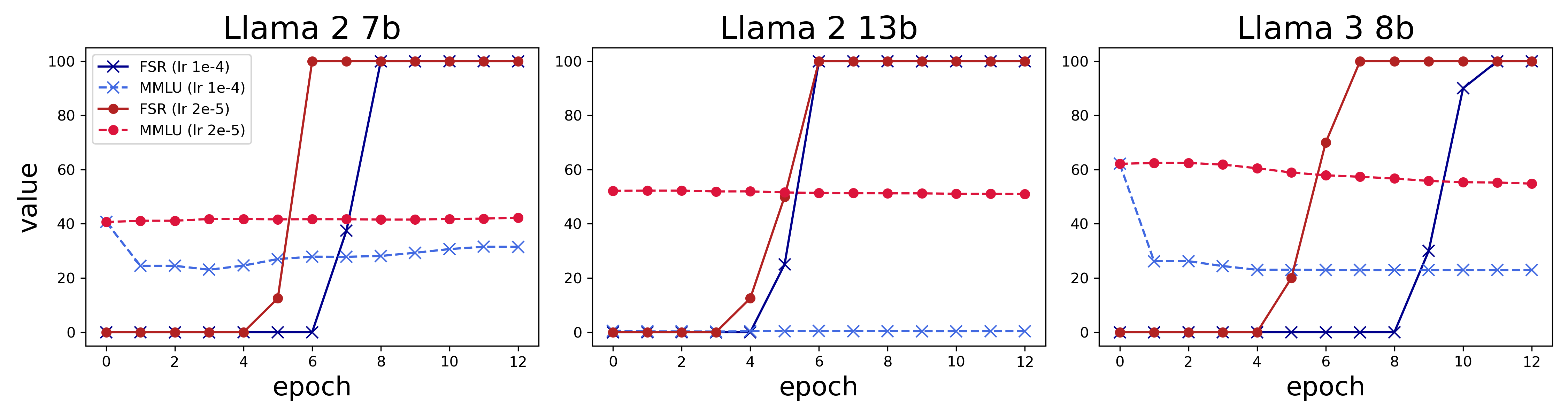}
\vspace{-1ex}
\caption{Harmlessness Analysis.}
\label{figs:harmless_analysis.png}
\end{figure*}


\subsection{Unforgeable Property (Secrecy)}
\label{appendix:unforge}

We consider two strategies that an adversary uses to forge fingerprints.

First, the adversary can generate random query-response pairs and hope that they coincide with a target model. We show that analytically this is extremely unlikely. Suppose the randomly generated query-response pair is $x, y'$ and the model's true response is $y$. Since there is only one valid $y$ for any given $x$, the probability for $y'$ to be the same as $y$ is $\frac{1}{\mathcal{\mathcal{D}}^{|y|}}$ if the mapping from $x$ to $y$ is sufficiently random, where the length of $y$ is $|y|$ and the domain size of each token in $y$ is $\mathcal{D}$. In our implementation of \sys, $\mathcal{D}$ is 2,000 and $|y|$ is 9, yielding a probability of $1.95e-30$.

Second, the adversary randomly generates a set of fake fingerprint pairs and retrain the existing model on them so that the fingerprints can be removed. We will show that it is highly unlikely that the original fingerprints will be completely erased from the existing model. Suppose the adversary generates $N$ query-response pairs $x, y'$. The probability that the adversary model erases all $m$ fingerprint prompts of length $|x|$ is $\left(\frac{N}{\mathcal{D}^{|x|}}\right)^{m}$. In our implementation of \sys, $\mathcal{D}$ is 2,000, $|x|$ is $32$, and $m$ is 10. When $N=1,000$, it yields a probability of $4.68e-1027$. Even to achieve a $1\%$ probability of erasing the fingerprints, the adversary would need to generate $N=2.71e105$ query-response pairs. Therefore, completely erasing the fingerprints is statistically impossible due to the astronomically large data volume required.

\subsection{Harmless Property}
\label{appendix:harmless}
In Fig.~\ref{figs:harmless_analysis.png}, we show the fingerprint success rate and MMLU score on IF-SFT baseline during the different learning epoch, starting from epoch 0 to epoch 12. While the fingerprint learning epoch increases, We found that MMLU scores drop on all three models, including Llama2 7B, 13B, and Llama 3 8B,indicating that model performance is harmed by the fingerprint learning with IF-SFT.
We also discover different learning rate will affect the MMLU performance.

\end{document}